\documentclass{article}

\usepackage[final,main]{neurips_2026}
 \usepackage{amsmath}
 \usepackage{float}
 \usepackage{seqsplit}
\usepackage{booktabs, multirow}
\usepackage{algpseudocode}
\usepackage{wrapfig}
\usepackage{graphicx}

\usepackage{microtype}                       
\usepackage[font=small,skip=4pt]{caption}
\usepackage{enumitem}\setlist{nosep,leftmargin=1.2em}
\usepackage[utf8]{inputenc} 
\usepackage[T1]{fontenc}    
\usepackage{hyperref}       
\usepackage{url}            
\usepackage{booktabs}       
\usepackage{amsfonts}       
\usepackage{nicefrac}       
\usepackage{microtype}      
\usepackage{xcolor}         
\usepackage{graphicx}
\usepackage{algorithm}

\newcommand{\chud}{\textsc{CHAD}}

\title{Probabilistic Steering of Generative Priors for Lab-in-the-Loop Antibody Optimization}

\workshoptitle{Probabilistic Steering of Generative Priors for Lab-in-the-Loop Antibody Optimization}

\author{%
  \textbf{Anisha Parsan$^*$} \\
  Massachusetts Institute of Technology \\
  \texttt{aparsan@mit.edu}
  \And
  \textbf{Sebastian Ober} \\
  BigHat Biosciences
  \And
  \textbf{Calvin McCarter} \\
  BigHat Biosciences
  \AND
  \\[-2em]
  \textbf{Nick Bhattacharya} \\
  BigHat Biosciences
  \And
    \\[-2em]
  \textbf{Hunter Elliott} \\
  BigHat Biosciences
}

\begin{document}

\maketitle
\begingroup
\renewcommand{\thefootnote}{*}
\footnotetext{Work done while at BigHat Biosciences.}
\endgroup

\begin{abstract}

Therapeutic antibody development is often framed as comprising two discrete stages: initial discovery of novel binders followed by iterative optimization to refine these hits into therapeutic-grade molecules. Models and methods often reflect these stages, with generative structure models applied to discovery and probabilistic black-box optimization methods applied to refinement. Here we propose to unify these stages by smoothly transitioning a generative discovery model into iterative optimization, allowing the handoff between the structure-based prior and supervision from wet lab data to be guided by the posterior uncertainty of a probabilistic surrogate. We show that this yields improvements in optimization efficiency in the low-$N$ regime, as well as improved diversity amongst the optimized molecules.

\end{abstract}

\section{Introduction}

Engineered antibodies are a critical therapeutic format in modern medicine. However, the vast sequence search space coupled with the low-throughput nature of experimental validation makes exhaustive search infeasible. Recent sequence-structure generative models now propose de-novo binders to specified epitopes without experimental measurements~\citep{bennett2025rfantibody, cho2025boltzdesign, stark2025boltzgen}. Yet these samples remain far from viable therapeutics: hit rates are low, affinities are often micromolar, and developability is rarely met~\citep{millefragoso2025germinal}.

Optimizing a de-novo hit is therefore treated as a separate problem, handed to a distinct stack of supervised models trained on lab-measured properties to select each experimental batch ~\citep{frey2025labinloop, stanton2022accelerating}. As wet-lab testing dominates the cost of each round, low $N$ efficiency -- fitness gain per variant measured -- is critical. While de-novo models excel at $N=0$, supervised surrogates become increasingly informative as measurements accumulate. Thus, low-$N$ optimization remains an open question. Ideally one model would span both regimes: a generative prior with a calibrated handoff to the surrogate as data accumulate.

We realize this through inference-time steering of a frozen generative model toward a reward-tilted distribution, upweighting high-reward designs while keeping samples on the generative manifold~\citep{singhal2025fk}. To achieve this, we employ Feynman-Kac (FK) steering, which resamples a population of particles during denoising in proportion to a potential evaluated on the predicted clean estimate of each particle~\citep{singhal2025fk}. As FK only evaluates the reward, arbitrary lab-derived signals can steer sampling~\citep{didi2026complexa, hartman2025fk}. Crucially, we propose a surrogate-based acquisition as the reward potential~\ref{app:related}, enabling the discovery-to-optimization handoff by inheriting the surrogate's uncertainty.

We propose $\chud$ (Calibrated Handoff via Acquisition-guided Diffusion) with these contributions:
\begin{itemize}
    \item We recast antibody optimization as inference-time particle steering of a diffusion-based prior.
    \item We demonstrate that steering with a probabilistic acquisition function-based reward preserves batch diversity and explores divergent regions of sequence space.
    \item We achieve low-$N$ efficiency over sequence-only baselines as well as an un-guided structure ablation in multiple \textit{in-silico} campaigns.
\end{itemize}

\section{Methods}

\paragraph{Background.}
\label{sec:retro}

An antibody sequence is a string $x = (x^{(1)}, \dots, x^{(L)}) \in \mathcal{V}^L$, with $|\mathcal{V}| = 20$ amino acids; the CDRs are a subset of positions $\mathcal{I} \subseteq \{1, \dots, L\}$ which confer most of the binding specificity. We hold framework residues fixed and optimize only $\mathcal{I}$. At round $r$, a batch of $B$ designs query an oracle $f$ – in the lab, an assay – which returns fitness measurements. These are added to dataset $\mathcal{D}_r$, and a surrogate $g_r$ is refit on $\mathcal{D}_r$ to approximate $f$. For each completed campaign, a fixed oracle $f$ trained on the full data replaces the lab as the evaluator; we replay the active-learning loop against it.



\subsection{FK-Steering Formulation}
\paragraph{Sampling from the reward-tilted distribution.}
Let $p_\theta(x)$ be the frozen generative prior and $a_r(x)$ the round-$r$ reward over designs. To construct batches enriched for high fitness, we seek to sample from the reward-tilted distribution $p_r(x) \propto p_\theta(x)\exp(\lambda a_r(x))$ \refstepcounter{equation}(\theequation)\label{eq:tilted}. Feynman--Kac (FK) steering approximates draws from \eqref{eq:tilted} via sequential Monte Carlo \citep{singhal2025fk}. We carry a population of $P$ particles -- partially denoised states advanced in parallel under $p_\theta$ -- and, at a fixed set of intermediate denoising steps, resample them in proportion to a potential $G_r(x) = \exp\!\big(\lambda\,a_r(x)\big)$ evaluated on each particle's current clean estimate $\hat{x}_1$ (defined in \eqref{eq:x1hat}). Here the reward $a_r$ is the surrogate acquisition (Section~\ref{sec:reward}) and $\lambda$ is the steering strength. At a resample step, particle $i$ receives the normalized weight
\begin{equation}
w_i \;=\; \frac{G_r(\hat{x}_1^{(i)})}{\sum_{j=1}^{P} G_r(\hat{x}_1^{(j)})} \;=\; \frac{\exp\!\big(\lambda\,a_r(\hat{x}_1^{(i)})\big)}{\sum_{j=1}^{P}\exp\!\big(\lambda\,a_r(\hat{x}_1^{(j)})\big)},
\label{eq:fk-weights}
\end{equation}
and the next generation is drawn by sampling $P$ indices from $\mathrm{Multinomial}(w_1,\ldots,w_P)$ with replacement; surviving particles then continue denoising. We adopt the \emph{immediate} potential \citep{hartman2025fk}, which evaluates $G_r$ freshly on the current clean estimate at each resample step, contrasting potentials that accumulate reward across the trajectory \citet{singhal2025fk, wu2023practical}.

\paragraph{Implementation in Sequence--Structure Flow Matching.}
We use Proteina-Complexa \citep{didi2026complexa}, a target-conditioned flow-matching binder generator, as our prior $p_\theta$, and advance the $P$ particles of \eqref{eq:fk-weights} through it in parallel. Each per-residue state pairs an $\alpha$-carbon coordinate with an $8$-dimensional latent that a frozen autoencoder decodes to amino-acid identity. A design is generated by integrating from Gaussian noise at $t{=}0$ to $t{=}1$ over $T = 400$ steps; at each step the predicted velocity $v_\theta(x_t, t)$ gives the clean estimate in closed form,
\begin{equation}
\hat{x}_1 \;=\; x_t + (1-t)\,v_\theta(x_t, t),
\label{eq:x1hat}
\end{equation}
after which the state is advanced by a stochastic sampling step with scheduled noise \citep{didi2026complexa}. To redesign only positions $\mathcal{I}$, the remaining residues are overwritten each step with the parent's ground-truth coordinate and latent. A warm-start fraction $\gamma \in [0,1]$ controls deviation from the seed: it is noised to step $\lfloor \gamma T \rfloor$ and denoised to completion, so $\gamma{=}0$ recovers de-novo while larger $\gamma$ bounds the expected edit distance (Fig. ~\ref{fig:edits_vs_start}). See Appendix~\ref{app:x1approx}.

\subsection{Uncertainty-guided Reward Functions}
\label{sec:reward}

\paragraph{Acquisition function as reward potential.}
At round $r$ we fit a Gaussian-process surrogate $g_r$ on all measurements $\mathcal{D}_r$, refit from scratch each round. Each design is represented by a one-hot encoding of $\mathcal{I}$, over which the GP uses a Tanimoto kernel. The surrogate yields, for any candidate $x$, a posterior predictive mean $\mu_r(x)$ and standard deviation $\sigma_r(x)$. We steer with the log probability-of-improvement (logPI) acquisition, used directly as the FK reward $a_r$ in the potential $G_r$ of \eqref{eq:fk-weights}:
\begin{equation}
a_r(x) \;=\; \log \Phi\!\big(z_r(x)\big), \qquad z_r(x) \;=\; \frac{\mu_r(x) - y^\star_r}{\sigma_r(x)},
\label{eq:logpi}
\end{equation}
where $\Phi$ is the standard normal CDF and $y^\star_r = \max_{x_i \in \mathcal{D}_r} \mu_r(x_i)$ is the incumbent, taken as the largest posterior mean over the measured designs.
\paragraph{Batch construction.}
\label{sec:batch}
From the pooled proposals we select the measured batch of size $B$ using qHSRI \citep{Binois2025, Ober2025}, which ranks candidates by a Sharpe ratio over the Pareto front of $\big(\mu_r(x), \sigma_r(x)\big)$. Additionally, we seed the design of the next batch from a set of incumbents from the previous round (Appendix~\ref{app:incumbents}).

\subsection{The $\chud$ Active Learning Method}
\label{sec:task}
Algorithm~\ref{alg:chud} assembles the full active-learning loop. Each round refits the GP on all measurements so far and reads out the logPI reward $a_r$, then builds a proposal pool by repeated steered sweeps (Algorithm~\ref{alg:fksweep}). The round's incumbents seed the next round. As $\sigma_r$ contracts with data, the loop hands off from prior-driven exploration to surrogate-driven exploitation without change to the procedure.

\begin{figure}[h]
\vspace{-10pt}
\centering
\begin{minipage}[t]{0.49\linewidth}
\begin{algorithm}[H]
\caption{$\chud$}
\label{alg:chud}
\small
\begin{algorithmic}[1]
\Require prior $p_\theta$; parent $x^{\text{par}}$; rounds $R$; sweeps $K$; batch size $B$; strength $\lambda$
\State $\mathcal{D}_0 \gets \{(x^{\text{par}}, f(x^{\text{par}}))\}$
\State $S_0 \gets \{x^{\text{par}}\}$
\For{$r = 1, \dots, R$}
  \State fit GP on $\mathcal{D}_{r-1}$
  \State $a_r \gets \log\text{PI}$ from $(\mu_r, \sigma_r)$
  \State $\mathcal{P}_r \gets \bigcup_{k=1}^{K} \textsc{FKSweep}(p_\theta, S_{r-1}, a_r, \lambda)$
  \State $\mathcal{B}_r \gets \textsc{qHSRI}(\mathcal{P}_r, \mu_r, \sigma_r, B)$
  \State $\mathcal{D}_r \gets \mathcal{D}_{r-1} \cup \{(x, f(x)) : x \in \mathcal{B}_r\}$
  \State $S_r \gets$ diverse incumbents from $\mathcal{B}_r$
\EndFor
\end{algorithmic}
\end{algorithm}
\end{minipage}
\hfill
\begin{minipage}[t]{0.49\linewidth}
\begin{algorithm}[H]
\caption{\textsc{FKSweep}: $P$ particles}
\label{alg:fksweep}
\small
\begin{algorithmic}[1]
\Require prior $p_\theta$; seeds $S$; reward function $a_r$; strength $\lambda$; warm start $\gamma$; steps $\mathcal{T}_{\text{rs}}$; particles $P$
\State init $P$ particles: noise $S$ to $t_0 = \lfloor \gamma T \rfloor$
\For{$t = t_0, \dots, T$}
  \State $\hat{x}^{(i)}_1 \gets x^{(i)}_t + (1-t) v_\theta(x^{(i)}_t, t)$
  \If{$t \in \mathcal{T}_{\text{rs}}$}
    \State resample $\propto \exp\!\big(\lambda a_r(\text{decode}(\hat{x}^{(i)}_1))\big)$
  \EndIf
  \State advance step; clamp non-CDR to $x^{\text{par}}$
\EndFor
\State \Return $\{\text{decode}(x^{(i)}_T)\}_{i=1}^P$
\end{algorithmic}
\end{algorithm}
\end{minipage}
\end{figure}

\section{Experiments}

We evaluate $\chud$ along three core axes. First, we use $\chud$ to propose design rounds across three diverse targets and antibody formats (Section~\ref{sec:exp-efficiency}). Second, we isolate the role of the acquisition-based reward (Section~\ref{sec:exp-reward}). Finally, we characterize the mutational exploration of our setup (Section~\ref{sec:exp-exploration}).

\paragraph{Task.}
We draw data from three completed campaigns and simulate the active-learning protocol (Section~\ref{sec:retro}). Although $\gamma = 0$ would recover de-novo generation, in-silico oracles are only reliable on sequences in their training distribution, meaning a de-novo design would require extrapolation. We therefore start each campaign from an in-distribution parental with $\gamma = 0.75$ (step $\lfloor \gamma T \rfloor = 300$ of $T = 400$) and resample at the late steps $\{320, 340, 360, 380\}$ on the way to $t{=}1$. Starting structures are predicted with Protenix \citep{Zhang2026}(Appendix~\ref{app:campaigns}).




\paragraph{Campaigns.}
We evaluate on three campaigns spanning binder formats, redesign scopes, and oracle types and availability of empirical structure. (Table~\ref{tab:campaigns}). \textit{Trastuzumab--HER2} is a Fab--antigen campaign; we score with the CNN binder classifier of \citet{calvinmlm}. \textit{scFv--SARS-CoV-2} redesigns the Ab-14 parental of \citet{Engelhart2022}, scored by a CNN affinity oracle. \textit{VHH--PT} redesigns a VHH against a proprietary target (PT), scored by an ensemble oracle trained on BLI kinetics. Because the oracle's training set is small, we reject designs too far in edit distance. Full details are in Appendix~\ref{app:campaigns}.

\paragraph{Baselines.}
We compare $\chud$ against three baselines. As an ablation within our own pipeline, Best-of-$N$ (BoN) is $\chud$ run at $T_{rs} = $ null, i.e. the identical FK setup with no resampling, so all $P$ particles denoise from the prior. As a sequence-only baseline we run the genetic algorithm (GA) portfolio method of \citet{Ober2025} (the \texttt{c-onehot-T} variant, dubbed OberBO). As a steering baseline we run LaMBO \citep{stanton2022accelerating}, warm-started from our BoN/FK seed set with Expected-Improvement acquisition, since it provides no prior of its own. Each baseline is allowed to run until its matched per-round budget is exhausted and is then downselected to a batch of size $B$; on rare occasions, a method fails to propose enough unique variants (details in Appendix~\ref{app:baselines}).


\subsection{$\chud$ achieves greater data efficiency at low-$N$ across diverse targets.}
\label{sec:exp-efficiency}

\begin{figure}[h]
    \centering
    \includegraphics[width=0.85\linewidth]{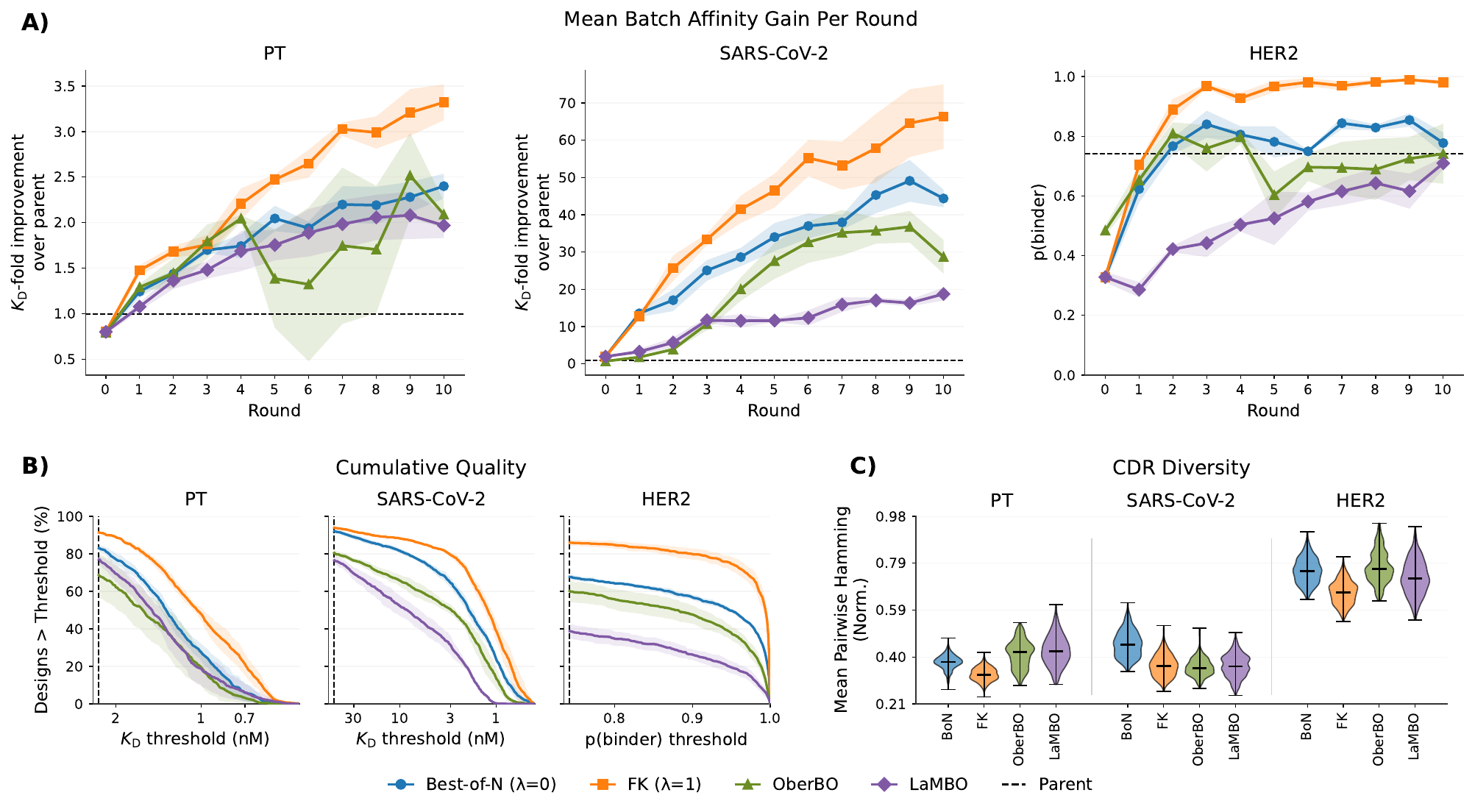}
    \caption{\textbf{(A)} Per-round fold-improvement over the parental on campaigns with predicted affinity; we directly plot \textit{p(binder)} from the HER2 campaign. \textbf{(B)} Quality vs. survival of all designs suggested across the full in-silico campaign. \textbf{(C)} Mean pairwise Hamming distance of sequences, normalized by number of mutable residues.}
   \label{fig:three_campaign_panel}

\end{figure}

\begin{wraptable}{r}{0.5\textwidth}
\vspace{-11pt}
\centering
\caption{\textbf{Design performance across three campaigns.} Best values per target are bolded.}
\label{tab:main}

\vspace{2pt}
\resizebox{0.5\textwidth}{!}{%
\begin{tabular}{llcccc}
\toprule
Target & Method & Designs & Best quality & Mean top-10 & $>$ parent (\%) \\
\midrule
\multirow{4}{*}{PT}
 & Best-of-$N$ & $1056$ & $4.1\times$ & $4.0\times$ & $83 \pm 3$ \\
 & FK & $1056$ & $\mathbf{5.1\times}$ & $\mathbf{4.8\times}$ & $\mathbf{91 \pm 1}$ \\
 & OberBO & $724^{*}$ & $4.0\times$ & $3.7\times$ & $69 \pm 20$ \\
 & LaMBO & $1003^{*}$ & $4.9\times$ & $4.6\times$ & $77 \pm 4$ \\
\midrule
\multirow{4}{*}{SARS-CoV-2}
 & Best-of-$N$ & $1056$ & $118.5\times$ & $110.7\times$ & $92 \pm 2$ \\
 & FK & $1056$ & $\mathbf{119.3\times}$ & $\mathbf{115.4\times}$ & $\mathbf{94 \pm 1}$ \\
 & OberBO & $1056$ & $100.7\times$ & $88.4\times$ & $80 \pm 3$ \\
 & LaMBO & $999^{*}$ & $60.7\times$ & $48.1\times$ & $77 \pm 4$ \\
\midrule
\multirow{4}{*}{HER2}
 & Best-of-$N$ & $1056$ & $1.000$ & $1.000$ & $68 \pm 3$ \\
 & FK & $1056$ & $1.000$ & $1.000$ & $\mathbf{86 \pm 2}$ \\
 & OberBO & $1056$ & $1.000$ & $1.000$ & $60 \pm 10$ \\
 & LaMBO & $991^{*}$ & $1.000$ & $1.000$ & $39 \pm 7$ \\
\bottomrule
\end{tabular}%
}

\vspace{2pt}
{\scriptsize $^{*}$Method did not reach $1{,}056$ unique designs under matched compute.}

\vspace{-8pt}
\end{wraptable}

\begin{figure}[h]
    \centering
    \includegraphics[width=0.85\linewidth]{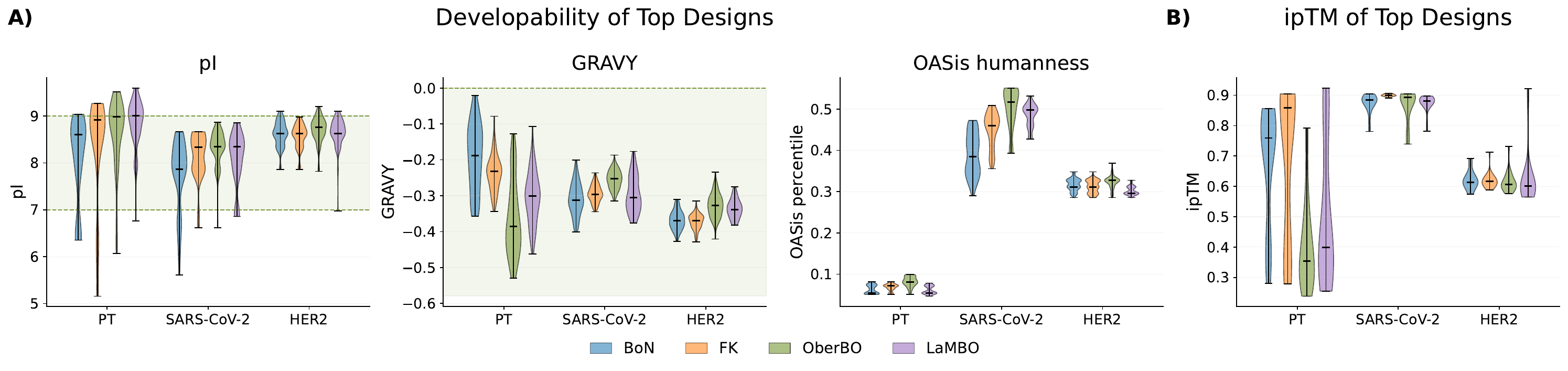}
    \caption{\textbf{(A)} Assessment of developability criteria across methods amongst the the top 10 \% of designs, with green indicating favored regions. \textbf{(B)} ipTMs of top 25 designs in each method folded with \textsc{Protenix V2}.}
   \label{fig:dev}
\end{figure}

We first investigate the data efficiency of $\chud$ (FK $\lambda{=}1$) against the baselines in Fig.~\ref{fig:three_campaign_panel}A. Across targets, $\chud$ outperforms baselines with mean batch fitness climbing faster in the early rounds and remaining high; on HER2, it reaches the $p(\text{binder})$ ceiling in the fewest rounds. Aggregated over the campaign (Table~\ref{tab:main}), $\chud$ yields both the best single design and the strongest mean top-$10$ on every target. In (Fig.~\ref{fig:three_campaign_panel}B), $\chud$ retains a higher fraction of designs above any fitness threshold across all three targets, and likewise in the fraction of designs that improve on the parent (Table~\ref{tab:main}). We note a limitation that $\chud$ does not achieve best mean CDR-diversity on any particular campaign, though diversity serves exploration and isn't an objective in itself (Fig.~\ref{fig:three_campaign_panel}C). Despite using a non-antibody-specific prior, $\chud$ roughly matches baselines across standard \textit{in-silico} developability metrics (Fig.~\ref{fig:dev}; Appendix Table~\ref{tab:dev}).

\subsection{Uncertainty-guided Rewards Suggest Diverse and High-Scoring Batches.}
\label{sec:exp-reward}

\begin{figure}[h]
    \centering
    \vspace{-8pt}
    \includegraphics[width=0.9\linewidth]{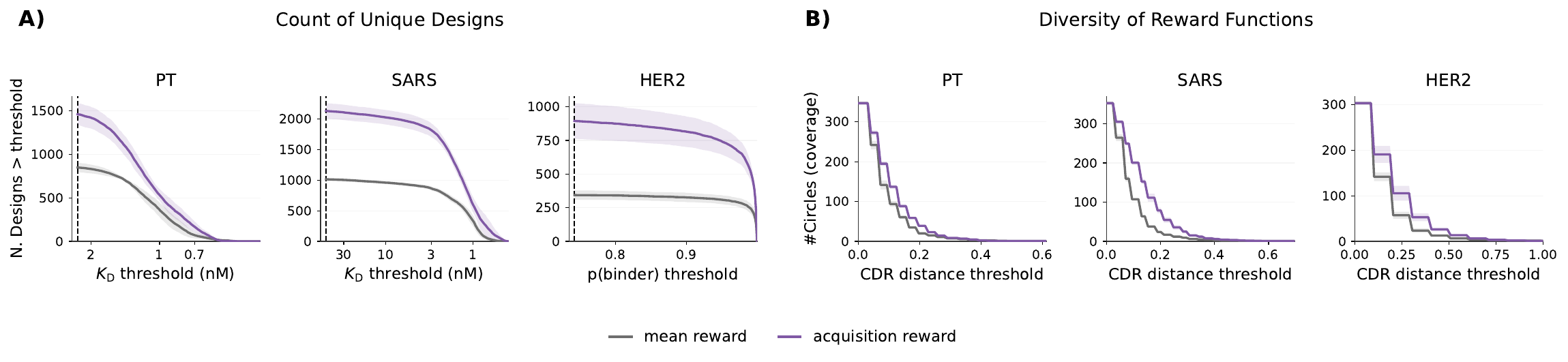}
    \caption{\textbf{(A)} Survival quality plots of all unique designs prior to batch selection. \textbf{(B)} \#Circles metric on the CDR distances within batches suggested by mean vs acquisition reward.}
   \label{fig:circles}
\end{figure}

To isolate the role of uncertainty, we steer on two rewards: the logPI acquisition ($\chud$) and the posterior mean $\mu_r$ alone. While achieving comparable performance across rounds, uncertainty guidance improves exploration of sequence space (Fig.~\ref{fig:mean_vs_aqn}). It yields roughly twice as many unique variants before batch selection, including at higher fitness values, suggesting that acquisition weighting avoids collapse onto a few high-scoring sequences, maintaining a broader population of strong candidates (Fig.~\ref{fig:circles}). This is useful when downstream developability filters or multi-objective constraints reduce the number of viable designs. We further quantify CDR-space coverage with \#Circles~\citep{xie2023how}, a packing-based diversity measure (Appendix~\ref{app:circles}). At matched sample size, acquisition steering covers more CDR space than mean steering, with AUC ratios of 
$>1.2\times$ across campaigns.

\subsection{Exploration of Sequence Space and Effects of Steering Controls}

We analyze whether the mutations $\chud$ proposes occupy a different part of sequence space than sequence-only baselines. The MDS embedding of pairwise CDR Hamming distances (Fig.~\ref{fig:MDS}) indicates that FK and BoN occupy a shared region that does not overlap substantially with the OberBO runs on 2 out of 3 targets. Such divergence is of practical value, as one often uses an ensemble of methods that explore distinct regions of sequence space. Examining mutations directly, we find that at positions mutated frequently by both methods, the identity of the substituted residue can differ substantially (Fig.~\ref{fig:mut_identity}, Fig.~\ref{fig:mut_scatter}). 
\label{sec:exp-exploration}

\begin{figure}[h]
    \vspace{-6pt}
    \centering
    \includegraphics[trim=0 0 0 10mm, clip,width=0.75\linewidth]{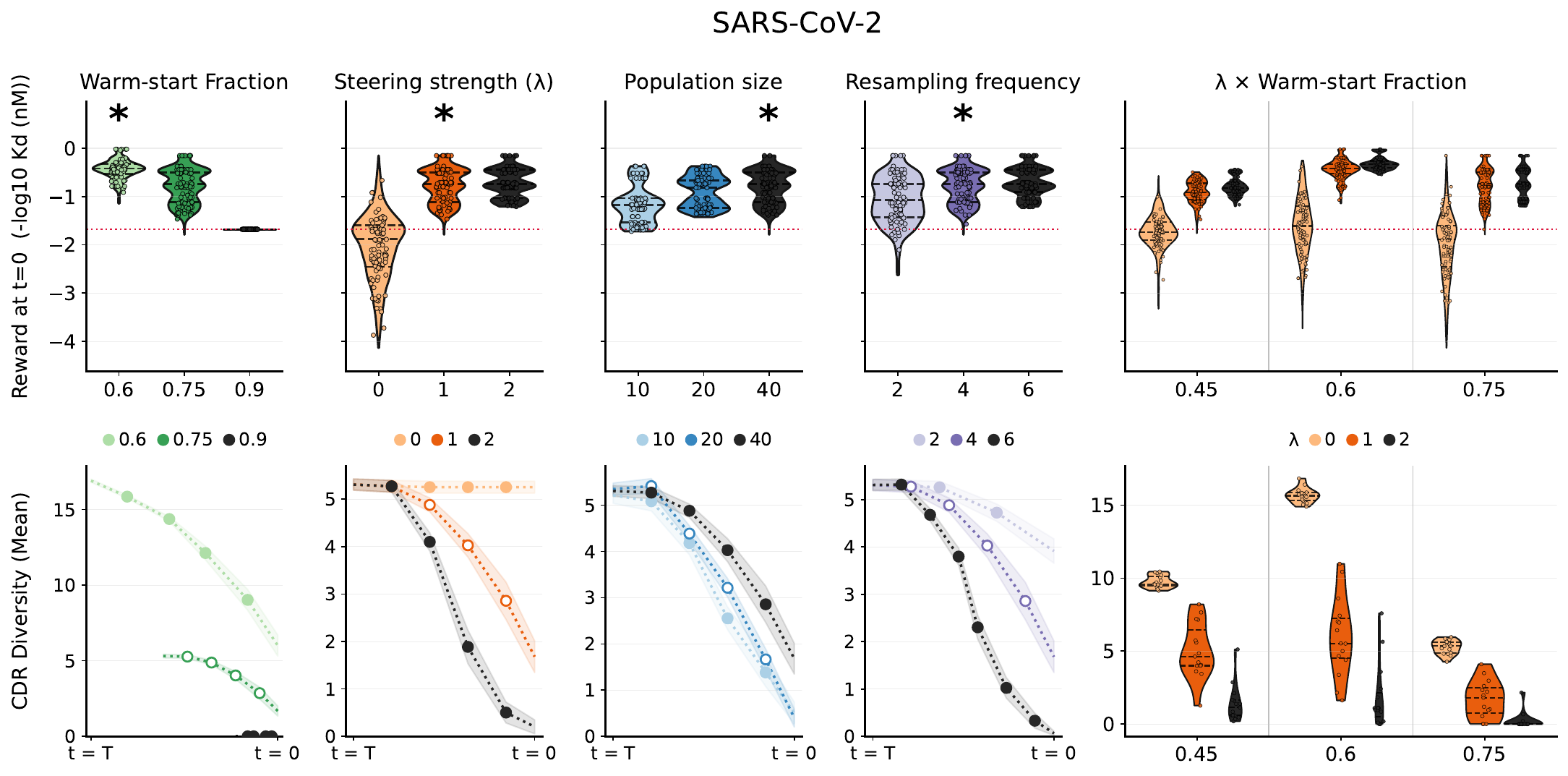}
    \caption{Effects of FK-steering presets on CDR diversity and final obtained reward on the SARS target. Experiments completed with one \ref{alg:fksweep} with the oracle $f$ in the loop to isolate effects of steering itself.}
   \label{fig:FK}
\end{figure}

 To isolate the effects steering itself, we run \textsc{FKSWEEP} with oracle $f$ directly as the reward. Lowering the warm-start fraction and raising the particle count both widen the pool of unique designs and increase diversity (Fig. \ref{fig:FK}). Contrarily, raising the steering strength $\lambda$ and resampling frequency concentrates the population on high-reward designs. Taken together, these controls offer several levers on the reward-diversity tradeoff (see App. \ref{app:restfk} for other targets).

\section{Discussion}

We demonstrate a method to transition generative priors to iterative optimization via steering, using a GP's acquisition function as the resampling potential. Across three diverse campaigns and formats, $\chud$ reaches higher fitness in fewer measured designs than baselines with no cost to general developability. Additionally, acquisition-guided steering surfaces roughly twice as many unique high scoring designs per round as steering on the posterior mean. 

Our in-silico evaluation has two key limitations. First, the seeds we optimize from are campaign-derived, whereas our method is intended to further-optimize de novo-discovered hits, a regime the retrospective setup cannot reproduce. Second, improvements are measured against learned in-silico oracles rather than wet-lab affinities, which may be exploitable outside their training distribution. Full validation therefore requires wet-lab experiments.

Future work should track $\sigma_r$ contraction over rounds to better characterize effects of uncertainty. Further, $\chud$'s scalar steering interface naturally extends to multi-objective optimization via Pareto-aware acquisition functions such as EHVI. Additionally, gradient-based steering could be explored with differentiable surrogates. Finally, because our prior is not antibody-specific, alternative generative architectures should be explored without changes to the formulation.

\bibliographystyle{plainnat} 
\bibliography{citations} 

\appendix

\section{Appendix}

\subsection{Related Work}
\label{app:related}

\paragraph{Inference-time steering of diffusion models.}
Gradient-based steering follows reward gradients through the denoiser's hidden states: in antibody optimization, LaMBO steers a discrete diffusion model with an acquisition objective~\citep{stanton2022accelerating, gruver2023lambo2}. This requires a differentiable reward and refitting the value head against the architecture each round, a cost that compounds as priors grow. Feynman--Kac (FK) steering instead resamples particles at intermediate denoising steps in proportion to a potential on each particle~\citep{singhal2025fk, hartman2025fk}.

\paragraph{Acquisition functions as guidance signals.}
Optimizing a weak hit is typically framed as supervised active learning, where an acquisition function trades a surrogate's predicted value against its uncertainty~\citep{frey2025labinloop, shahriari2016bo, stanton2022accelerating}.  Acquisition functions serve as guidance signals in reinforcement learning~\citep{auer2002ucb} and in proposals trained to sample proportional to a BO acquisition~\citep{brookes2019cbas, yun2025posterior}. In our work, we apply the acquisition as a resampling potential on a frozen prior, refit each round against lab measurements -- a setting prior FK work has not explored~\citep{hartman2025fk}.

\label{app:campaigns}

\subsection{Campaign Specifications}
Tables ~\ref{tab:campaigns} and ~\ref{tab:sequences} details the binder formats, sequence lengths, redesigned regions, and oracle evaluation criteria for each campaign. All starting structures were predicted using \texttt{protenix\_base\_default\_v1.0.0} with the default inference parameters (10 recycling cycles and 200 diffusion steps). For the Trastuzumab-HER2 and VHH-PT complexes, crystal structures of the bound complex were used as templates, and MSAs were used for all three complexes and computed via the ColabFold MSA server, \texttt{api.colabfold.com}. For the Trastuzumab–HER2 complex, we truncate the modeled system to the VH and VL domains of the Fab and to HER2 subdomain IV for tractability; the sequences used are given in Table 3.  Since the wild-type scores p(binder) $\approx 1.0$, we seed from a parental with $50\%$ of CDR-H3 randomized ($p(\text{binder}) \approx 0.74$) Additionally, because the PT oracle is trained on a comparatively small measured set, its predictions degrade outside the edit-distance range that set spans. For this campaign only, we therefore reject proposed designs more than 9 CDR edits from the oracle's training distribution prior to batch selection. The filter is applied identically to all arms.

\begin{table}[htbp]
\centering
\small
\caption{Campaign complexes and evaluation targets.}
\label{tab:campaigns}
\begin{tabular}{lllll}
\toprule
Campaign & Binder format (length) & Antigen & Redesigned (\#) & Target \\
\midrule
VHH--PT
  & VHH (117\,aa)
  & PT (internal)
  & 3 CDRs (31)
  & $-\log_{10}(K_D $ in $ \mathrm{M})$ \\
scFv--SARS-CoV-2
  & scFv (246\,aa)
  & SARS-CoV-2 peptide
  & $V_L$ CDRs (33)
  & $\log_{10}(K_D $ in $ \mathrm{nM})$ \\
Trastuzumab--HER2
  & Fv (227\,aa)
  & HER2 subdomain IV
  & CDR-H3 (10)
  & $p(\mathrm{binder})$ \\
\bottomrule
\end{tabular}
\end{table}

\begin{table}[h]\centering\scriptsize
\caption{Public-campaign complexes: parent binder, antigen, and redesigned CDR ranges (0-indexed).}
\label{tab:sequences}
\begin{tabular}{@{}lp{5.2cm}p{2.4cm}l@{}}
\toprule
complex & binder\_seq & target\_seq & redesign (0-idx) \\
\midrule
\texttt{scfv\_sars} &
{\ttfamily\seqsplit{EVQLVETGGGLVQPGGSLRLSCAASGFTLNSYGISWVRQAPGKGPEWVSVIYSDGRRTFYGDSVKGRFTISRDTSTNTVYLQMNSLRVEDTAVYYCAKGRAAGTFDSWGQGTLVTVSSGGGGSGGGGSGGGGSDVVMTQSPESLAVSLGERATISCKSSQSVLYESRNKNSVAWYQQKAGQPPKLLIYWASTRESGVPDRFSGSGSGTDFTLTISSLQAEDAAVYYCQQYHRLPLSFGGGTKVEIK}} &
{\ttfamily\seqsplit{PDVDLGDISGINAS}} &
156--172; 188--194; 227--235 \\
\addlinespace
\texttt{trastuzumab\_her2\_h3mut} &
{\ttfamily\seqsplit{EVQLVESGGGLVQPGGSLRLSCAASGFNIKDTYIHWVRQAPGKGLEWVARIYPTNGYTRYADSVKGRFTISADTSKNTAYLQMNSLRAEDTAVYYCSRFEGDWFDAMYYWGQGTLVTVSSDIQMTQSPSSLSASVGDRVTITCRASQDVNTAVAWYQQKPGKAPKLLIYSASFLYSGVPSRFSGSRSGTDFTLTISSLQPEDFATYYCQQHYTTPPTFGQGTKVEIK}} &
{\ttfamily\seqsplit{HPECQPQNGSVTCFGPEADQCVACAHYKDPPFCVARCPSIWKFPDEEGACQPCPIN}} &
98--107 \\
\bottomrule
\end{tabular}
\end{table}

\subsection{\textit{In-silico} Affinity Oracles.} 

\paragraph{The internal program oracle.} The affinity oracle we used is an ensemble of 10 CARP/ByteNet regressors with 32 residual blocks \citep{yang2024convolutions}, 
trained using cross-validation with distinct 0.8/0.1/0.1 fractional train/validation/test splits.
We pretrained the models on approximately 658,000 sequences from a phage display affinity maturation campaign, processed using Next
Generation Sequencing (NGS). 
We subsequently fine-tuned the ensemble on 1335 $K_D$ measurements derived from Bio-Layer Interferometry (BLI), performed on antibodies synthesized via
Cell Free Protein Synthesis (CFPS).
The resulting ensemble achieved a test cross-validated Spearman correlation of 0.71.

\paragraph{The SARS-CoV-2 affinity oracle.} We train a residual dilated 1-D CNN over the VL. Per-residue one-hot input features are concatenated with $64$-dim PCA-reduced AbLang2 embeddings \citep{Olsen2022}, projected to $128$ channels by a $1{\times}1$ convolution, passed through four residual blocks of two dilated convolutions each (kernel $3$; dilations $1,2,4,8$), pooled by concatenated global mean and max ($\to 256$), and read out by a $256\!\to\!256\!\to\!64\!\to\!1$ MLP (ReLU, dropout $0.1$). We train with MSE on the standardized target, Adam (\texttt{lr = 1e-3, weight decay 1e-5, batch 256}) with early stopping on validation MAE. We train on both of the datasets in \cite{Engelhart2022} filtered to the Ab-14 parental heavy-chain rows and averaged per unique VL (59,019 unique VLs, a random 80/10/10 split (47,217 / 5,901 / 5,901). We assess the fidelity of the model on the validation set below: 

\begin{table}[H]
\centering
\small
\caption{Deployed SARS-CoV-2 affinity oracle on the held-out test set, bucketed by CDR edit distance to the parent. Reported values are $n$-weighted averages. 
Overall ($n{=}5{,}901$): MAE $0.557$, Spearman $0.858$.}
\label{tab:scfv-oracle-distance}
\begin{tabular}{lrrrrr}
\toprule
Edit distance & d1--3 & d4--9 & d10--14 & d15--19 & d20--26 \\
\midrule
$n$      & 2{,}475 & 1{,}765 & 528   & 633   & 500   \\
MAE      & 0.794   & 0.335   & 0.430 & 0.438 & 0.457 \\
Spearman & 0.738   & 0.669   & 0.724 & 0.682 & 0.687 \\
\bottomrule
\end{tabular}
\end{table}

\paragraph{The Trastuzumab--HER2 affinity oracle.} We use the weights of the 1-D CNN classifier trained in of \citet{calvinmlm}. The model was trained over the one-hot $10$-mer CDR-H3 loops in the affinity-class data from \citet{Chinery2024} ($367{,}042$ Trastuzumab CDR-H3 variants labeled binder / non binder), which attains a held-out ROC-AUC of $0.995$ and PR-AUC of $0.990$. The sigmoid output probability saturates near $1.0$ for strong binders. Since the WT complex (PDB ID \texttt{1N8Z}) scores $\approx 1.0$, we seed from a deliberately weakened parent ($50\%$ of CDR-H3 randomized, $p(\text{binder})\approx 0.74$).

\subsection{Seeding with Incumbents}
\label{app:incumbents}
In our active learning loop, we seed the next ruond with incumbents from the previous round. We isolate the top $q$ fraction of designs by predicted value ($q = 0.2$) and greedily choose a maximally distant set by CDR Hamming distance.

\subsection{Baseline Details}
\label{app:baselines}

\paragraph{LaMBO} LaMBO-v1 \citep{stanton2022accelerating} optimizes in the latent space of a masked-language-model denoising autoencoder: each round a multi-task exact-GP surrogate is fit on the standardized measurements, and a single-objective genetic algorithm searches latent perturbations under an Expected-Improvement acquisition, decoding proposals back to sequences. Because LaMBO has no generative prior from which to draw a starting set, we warm-start each seed replicate from the same round-0 seed set used by our BoN/FK runs and match its budget to the other arms (32 seeds followed by 10 rounds of 32 measured designs). We retain all repository defaults, changing only the acquisition to Expected Improvement and the number of GA generations, which we set per target from per-generation timing probes so that LaMBO's per-round wall-clock matches the FK arms.

\paragraph{OberBO} OberBO applies the GA-portfolio BO stack of \citet{Ober2025} verbatim as a sequence-only baseline: at each round an NSGA-II genetic algorithm evolves a population of CDR-edit candidates that are ranked by qHSRI. We use the soft-constraint \texttt{c-onehot-T} preset (the one-hot Tanimoto kernel with the naturalness term), hold the GA presets at their defaults, and vary only the number of generations to match per-round compute; the GA is further restricted to at most three CDR edits per round relative to the running measured set, matching the $\approx$ 3 mutations per round induced by our warm-start fraction. The seed set is a 32-design cold start: the parent CDR sequence plus 31 unique variants, each formed by masking three randomly chosen CDR positions and multinomial-sampling residues from the masked-LM conditional, with a $4\times$ loop of oversampling and deduplicating until 32 unique designs are reached.

\subsection{Approximation of the denoised endpoint by $\hat{x}_1$}
\label{app:x1approx}

We resample at a fixed set of late steps, scoring each particle on $g_r$ via to the sequence obtained by decoding the particle's current clean estimate $\hat{x}_1$. While \citep{didi2026complexa} completes a full rollout to obtain the denoised sequence, we opt for $\hat{x}_1$ directly for computational efficiency; because we resample only at late steps, the resulting approximation error is small.

Specifically, we score each particle on its clean estimate $\hat{x}_1 = x_t + (1-t)\,v_\theta(x_t, t)$ rather than on a full denoising rollout to $t{=}1$. As $t \to 1$ the correction $(1-t)\,v_\theta$ vanishes, so $\hat{x}_1$ approaches the sampled endpoint $x_{\text{final}}$. Because our resampling steps $\mathcal{T}_{\text{rs}}$ lie in the late portion of the schedule, the error $\hat{x}_1 - x_{\text{final}}$ is small at every step where the reward is evaluated. 

\subsection{Circles Metric}
\label{app:circles}

We measure how broadly each arm covers sequence space with \#Circles~\citep{xie2023how}, the packing number of a set at threshold $t$: the largest subset in which every pair is more than $t$ apart. Higher is broader coverage; we report the curve over $t$ and summarize it by area under the curve (AUC).

First, we define distance over sequences rather than molecular fingerprints, using normalized Hamming distance. We restrict each design to its redesignable CDR positions. Third, while \citet{xie2023how} tunes a $t$ against a gold standard, we report coverage across the full range. Because packing counts grow with set size and the two arms propose different numbers of designs, we subsample every run to the smallest run in the campaign and average over 15 draws before counting. Across campaigns, acquisition/mean AUC ratios range from $1.20$ to $1.51$.

\subsection{Additional Results}

\begin{figure}[H]
    \centering
    \includegraphics[width=0.7\linewidth]{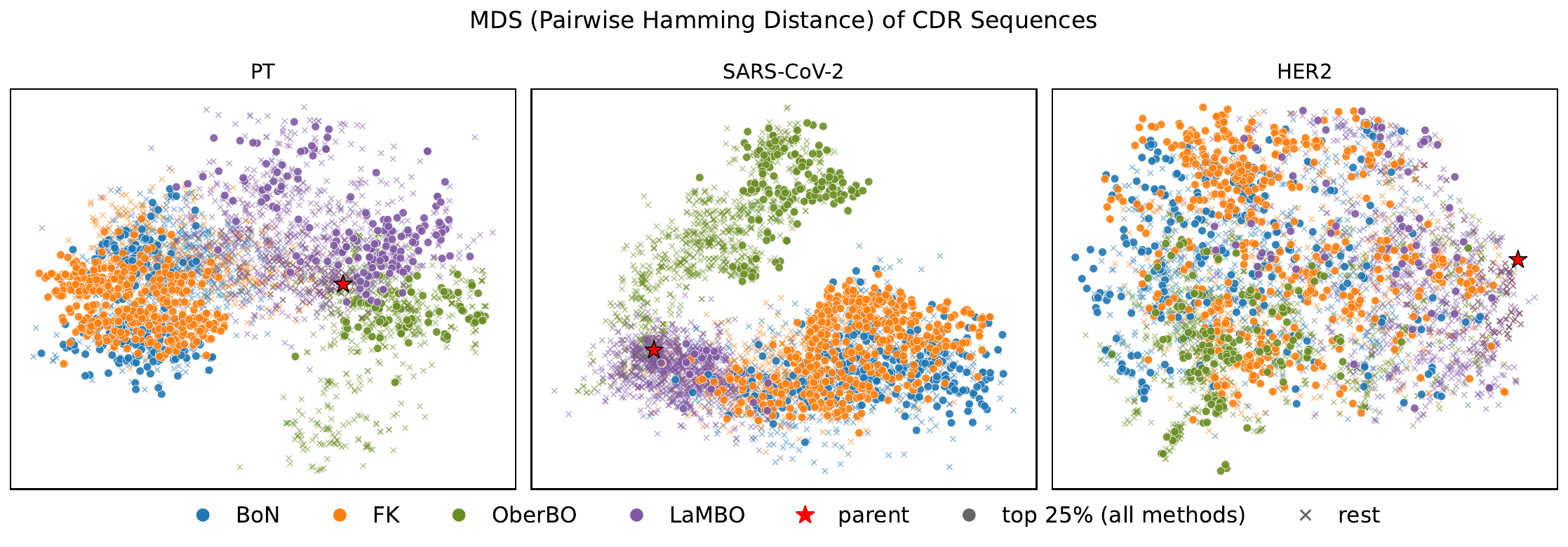}
    \caption{\textbf{MDS of Pairwise Hamming Distances Between CDRs of Designs.}. }
   \label{fig:MDS}
\end{figure}

\begin{table}[H]
\centering
\footnotesize
\setlength{\tabcolsep}{3pt}
\renewcommand{\arraystretch}{0.9}
\caption{Developability of proposed designs across three targets. Metrics are mean $\pm$ s.e.\
over 3 seeds.}
\label{tab:dev}
\begin{tabular}{llccc}
\toprule
Campaign & Method & \% pI $\in [7,9]$ & \% GRAVY $<0$ & OASis pct.\ (mean) \\
\midrule
\multirow{4}{*}{PT}
 & BoN        & $\mathbf{66 \pm 36}$ & $100 \pm 0$ & $0.062 \pm 0.010$ \\
 & FK         & $60 \pm 31$          & $100 \pm 0$ & $0.069 \pm 0.007$ \\
 & OberBO & $64 \pm 24$          & $100 \pm 0$ & $\mathbf{0.083 \pm 0.008}$ \\
 & LaMBO      & $48 \pm 10$          & $100 \pm 0$ & $0.062 \pm 0.006$ \\
\midrule
\multirow{4}{*}{SARS-CoV-2}
 & BoN        & $78 \pm 17$          & $100 \pm 0$ & $0.390 \pm 0.045$ \\
 & FK         & $93 \pm 8$           & $100 \pm 0$ & $0.451 \pm 0.032$ \\
 & OberBO & $\mathbf{98 \pm 3}$  & $100 \pm 0$ & $\mathbf{0.502 \pm 0.045}$ \\
 & LaMBO      & $82 \pm 8$           & $100 \pm 0$ & $0.483 \pm 0.029$ \\
\midrule
\multirow{4}{*}{HER2}
 & BoN        & $99 \pm 2$           & $100 \pm 0$ & $0.317 \pm 0.006$ \\
 & FK         & $\mathbf{100 \pm 0}$ & $100 \pm 0$ & $0.312 \pm 0.015$ \\
 & OberBO & $91 \pm 7$           & $100 \pm 0$ & $\mathbf{0.323 \pm 0.010}$ \\
 & LaMBO      & $97 \pm 0$           & $100 \pm 0$ & $0.303 \pm 0.004$ \\
\bottomrule
\end{tabular}
\end{table}

\begin{figure}[H]
    \centering
    \includegraphics[width=\linewidth]{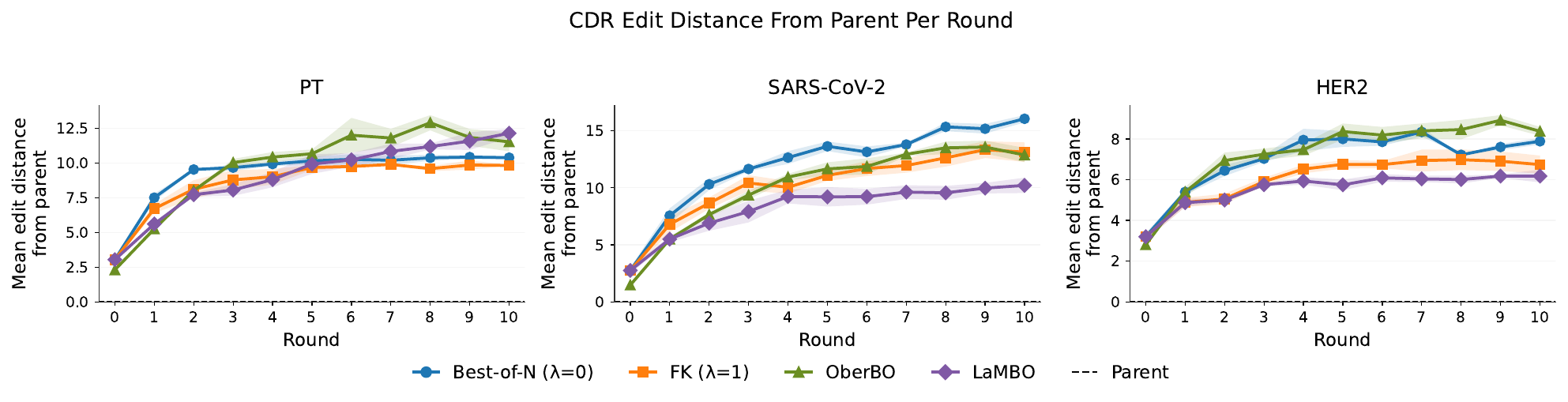}
    \caption{\textbf{CDR-Edits From Parental Per Round}. Average edit distance of proposed batches from the starting parental seed, ± S.E.}
   \label{fig:edits_from_parental}
\end{figure}

\begin{figure}[H]
    \centering
    \includegraphics[width=\linewidth]{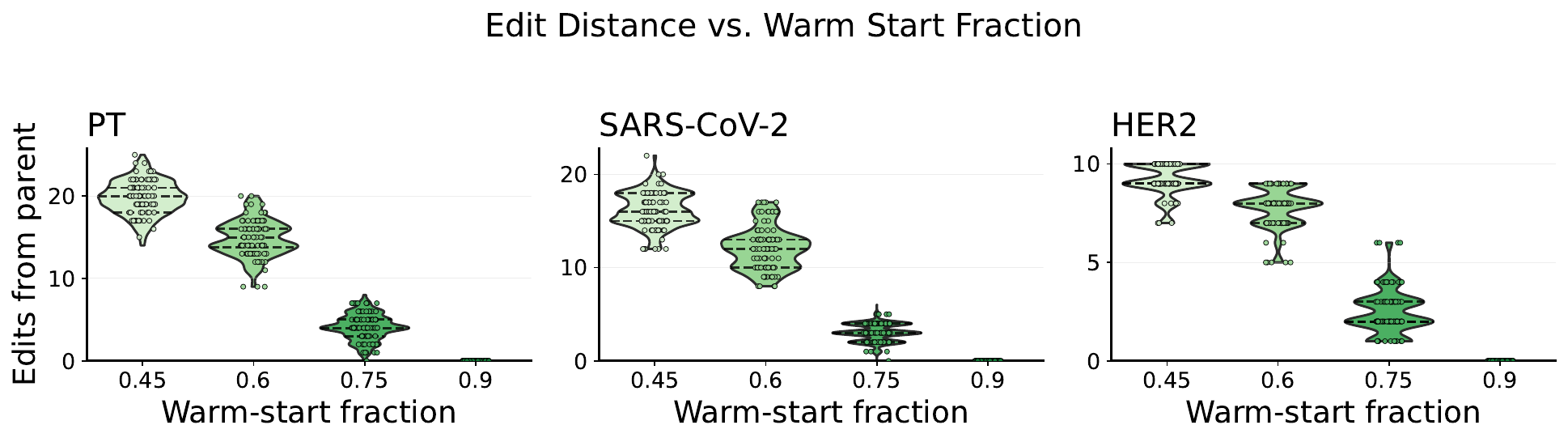}
    \caption{\textbf{Edit Distance vs. Warm Start Fraction}. Edit distance from the parental sequence achieved in designable regions at varying warm-start fractions }
   \label{fig:edits_vs_start}
\end{figure}

\begin{figure}[H]
    \centering
    \includegraphics[width=\linewidth]{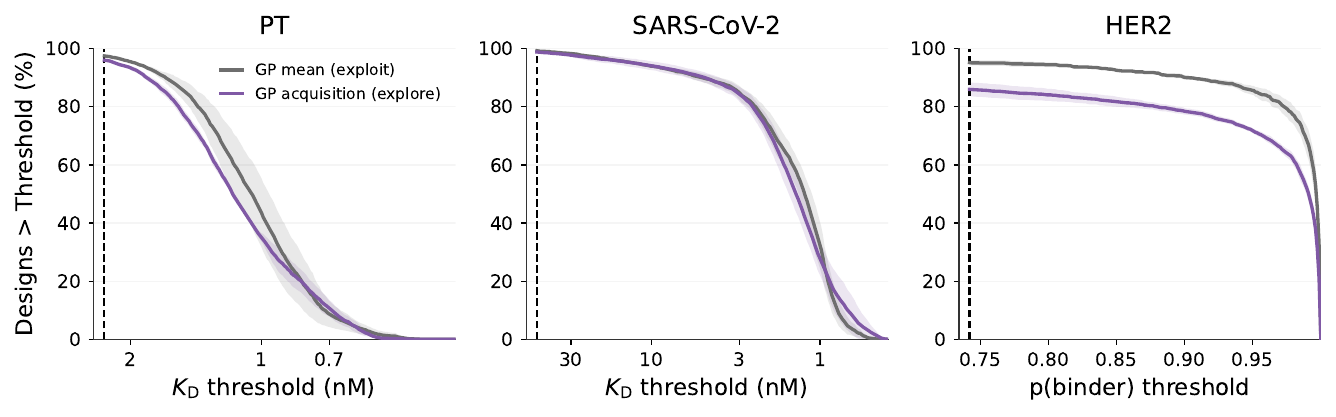}
    \caption{\textbf{Mean vs. Aquisition}. Survival quality curves in steering with acquisition reward vs. mean-only reward, ± S.E.}
   \label{fig:mean_vs_aqn}
\end{figure}

\begin{figure}[H]
    \centering
    \includegraphics[width=0.7\linewidth]{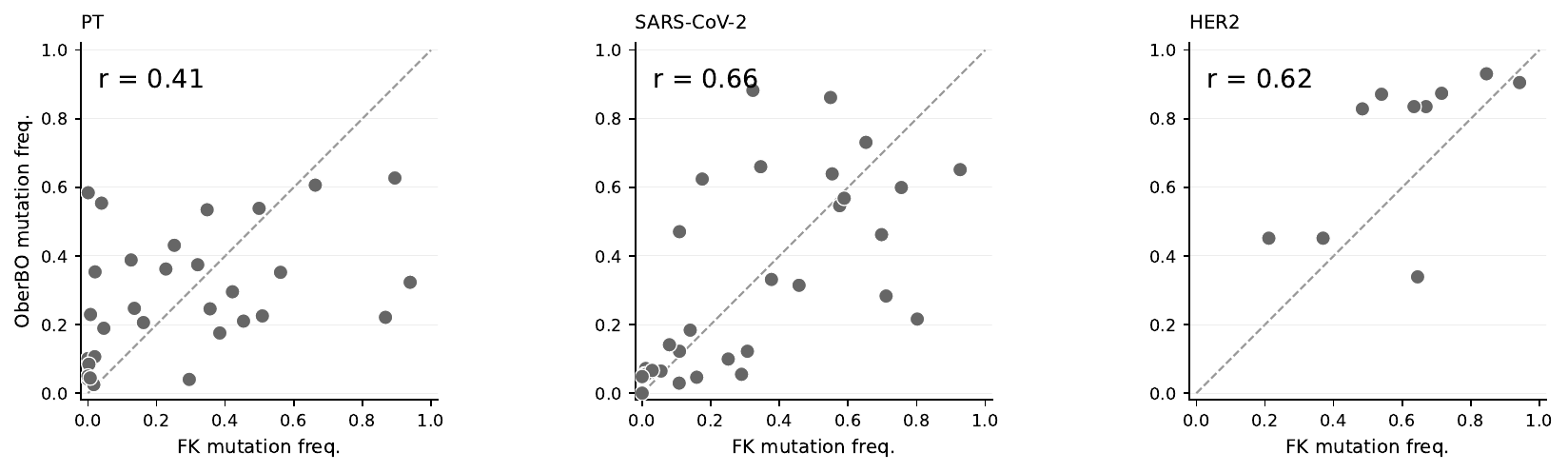}
    \caption{\textbf{Scatter of mutations across methods.} Scatterplot of mutation frequencies of each residue across methods (FK-steering and OberBO).}
   \label{fig:mut_scatter}
\end{figure}

\begin{figure}[H]
    \centering
    \includegraphics[width=0.7\linewidth]{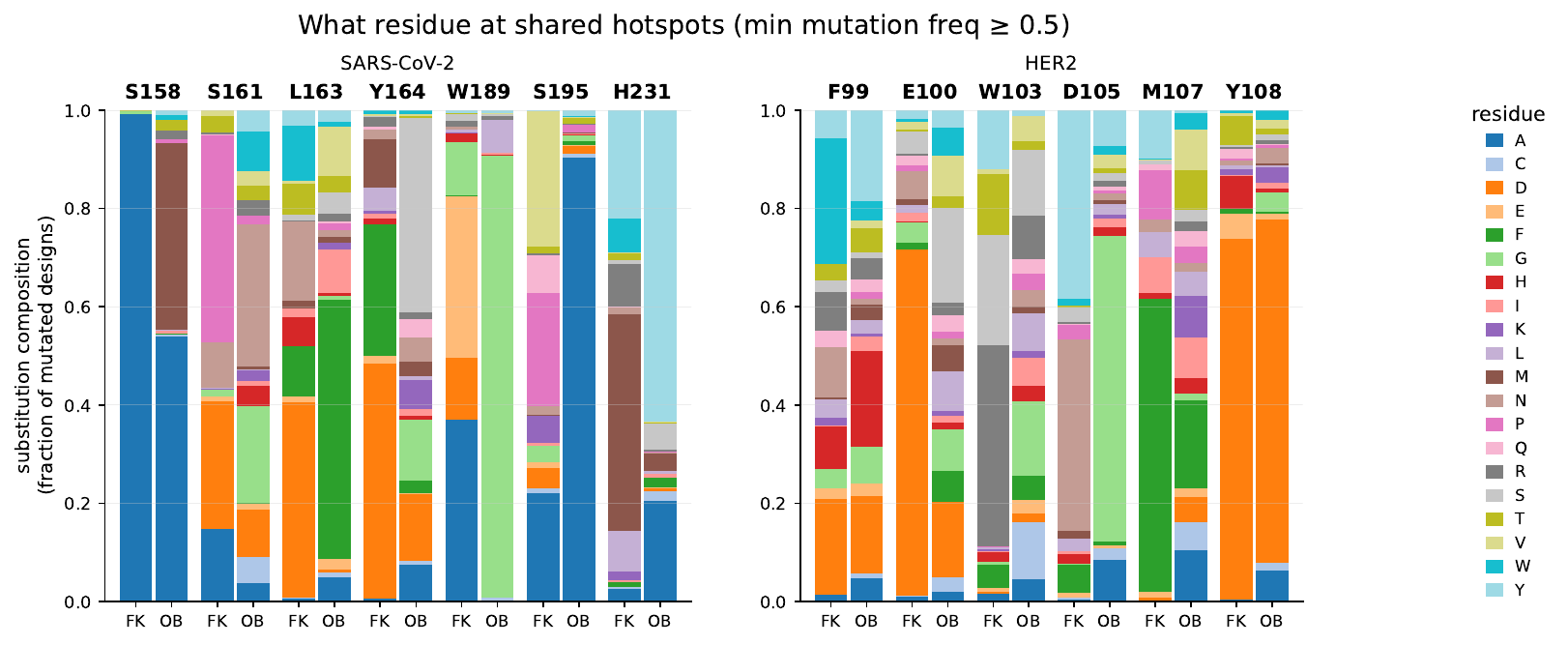}
    \caption{\textbf{Identity of amino acid mutations}. Breakdown of amino-acid substitution identity at shared mutational hotspots across methods (FK-steering and OberBO). }
   \label{fig:mut_identity}
\end{figure}

\begin{figure}[h]
    \centering
    \includegraphics[width=\linewidth]{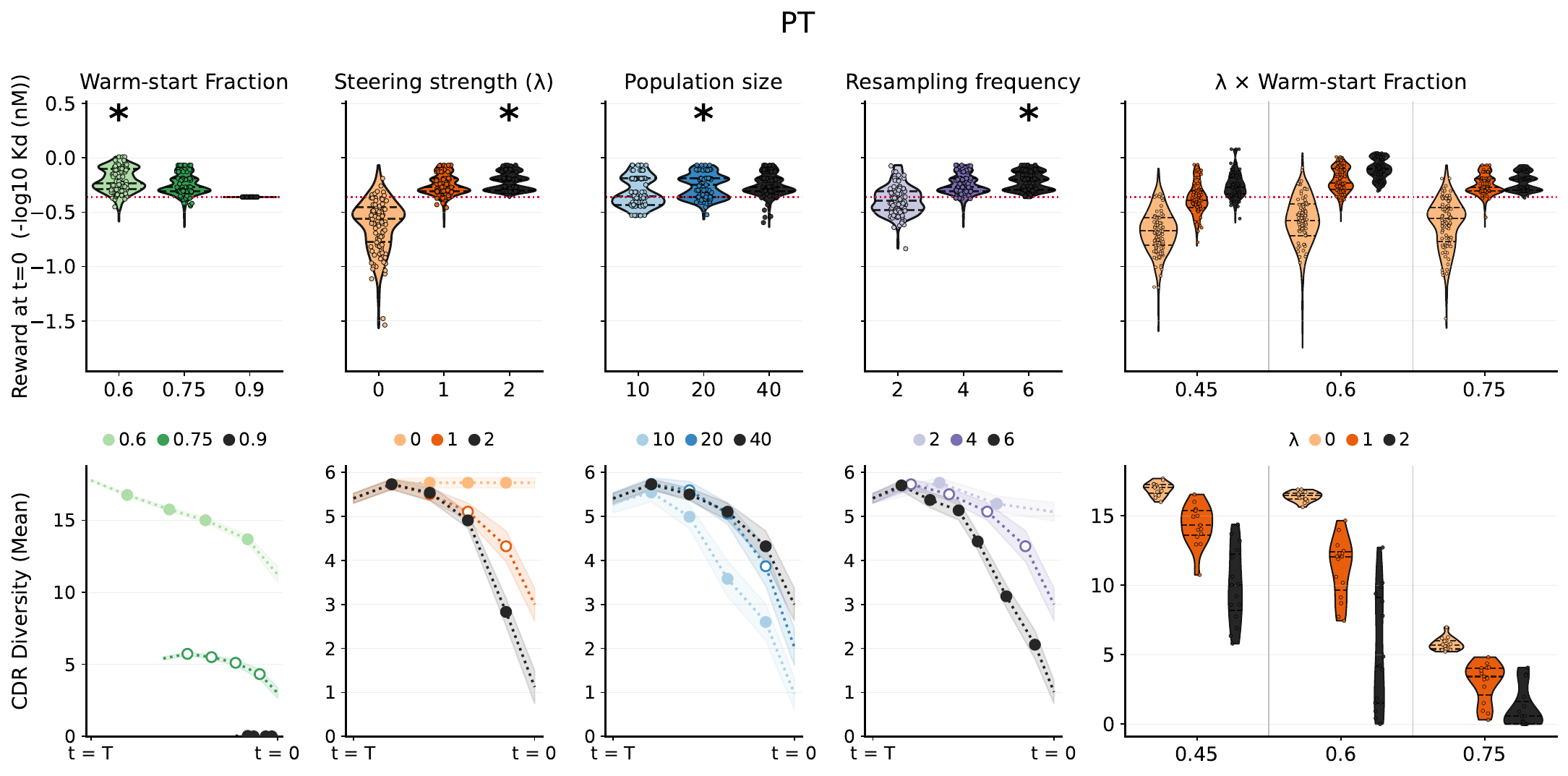}
    \caption{\textbf{PT Target: Effects of FK-steering Presets on CDR Diversity and Final Reward}}
   \label{app:restfk}
\end{figure}

\begin{figure}[H]
    \centering
    \includegraphics[width=\linewidth]{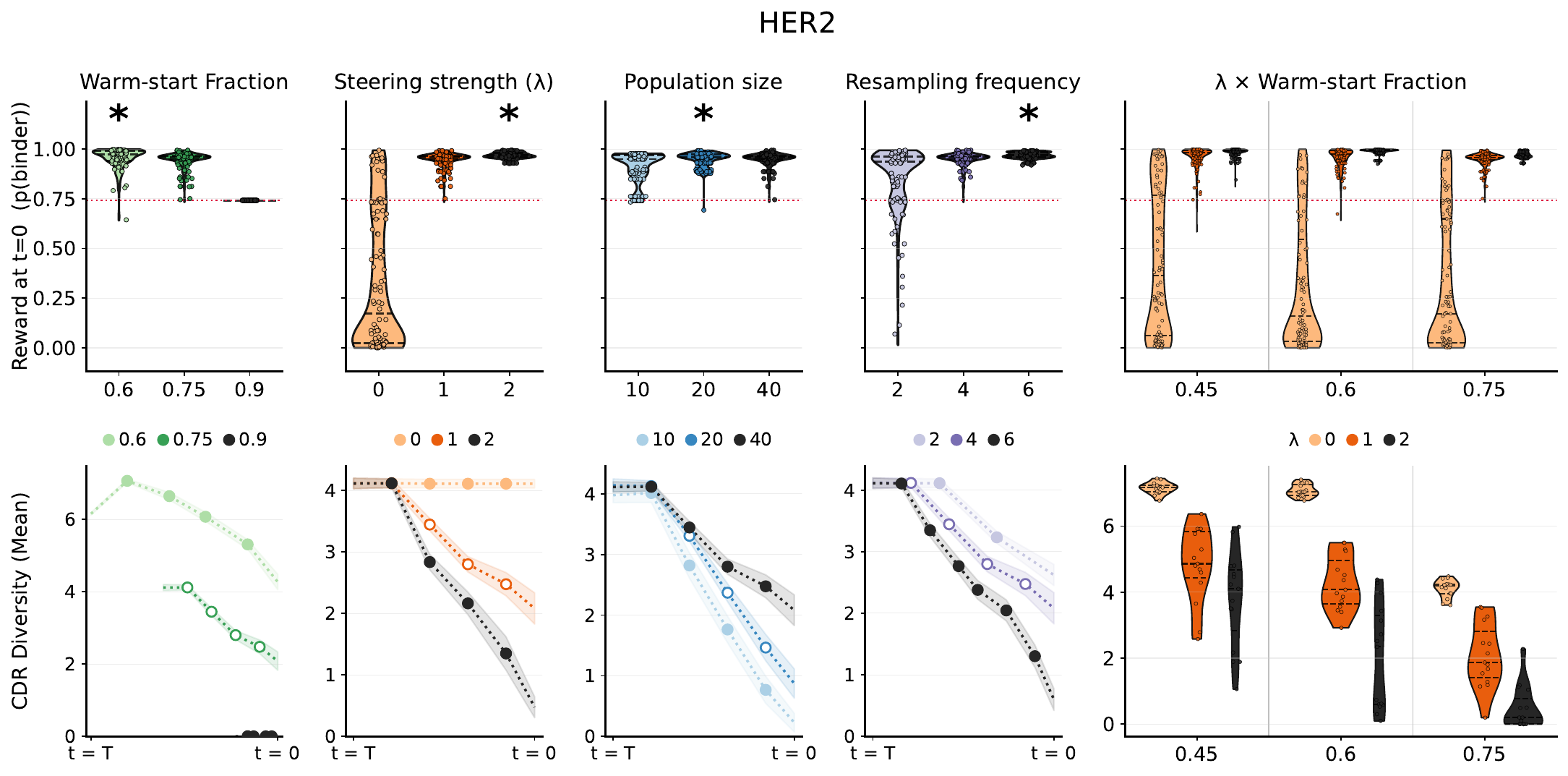}
    \caption{\textbf{HER2 Target: Effects of FK-steering Presets on CDR Diversity and Final Reward}}
   \label{app:restfk2}
\end{figure}



\end{document}